\documentclass[runningheads]{llncs}
\usepackage{graphicx}
\usepackage[acronym]{glossaries}
\usepackage{dsfont}
\usepackage{mathtools}
\usepackage{amsmath}
\usepackage{amssymb}
\usepackage{xcolor}
\usepackage{subfig}

\newcommand{\todo}[1]{}
\renewcommand{\todo}[1]{{\color{red} TODO: {#1}}}
\newcommand{\comment}[1]{}

\begin{document}

\newacronym{btcvae}{$\beta$-TCVAE}{$\beta$-Total Correlation Variational AautoEncoder}
\newacronym{bvae}{$\beta$-VAE}{$\beta$-Variational Autoencoder}
\newacronym{vae}{VAE}{Variational Autoencoder}

%
%\title{Towards disentangling feature representations in medical applications}
\title{Learning Interpretable Disentangled Representations using Adversarial VAEs}
%
%\titlerunning{Abbreviated paper title}
% If the paper title is too long for the running head, you can set
% an abbreviated paper title here
%
\author{Mhd Hasan Sarhan\inst{1,2}\and
Abouzar Eslami\inst{2} \and
Nassir Navab\inst{1,3} \and \\
Shadi Albarqouni\inst{1}}

\authorrunning{M.H Sarhan et al.}
% First names are abbreviated in the running head.
% If there are more than two authors, 'et al.' is used.
%
\institute{Computer Aided Medical Procedures (CAMP), Technische Universit\"at M\"unchen, Munich, Germany \and 
Carl Zeiss Meditec AG, Munich, Germany \and
Whiting School of Engineering, Johns Hopkins University, Baltimore, USA}
\maketitle              % typeset the header of the contribution
\begin{abstract}
Learning Interpretable representation in medical applications is becoming essential for adopting data-driven models into clinical practice. It has been recently shown that learning a disentangled feature representation is important for a more compact and explainable representation of the data. In this paper, we introduce a novel adversarial variational autoencoder with a total correlation constraint to enforce independence on the latent representation while preserving the reconstruction fidelity. Our proposed method is validated on a publicly available dataset showing that the learned disentangled representation is not only interpretable, but also superior to the state-of-the-art methods. We report a relative improvement of $81.50\%$ in terms of disentanglement, $11.60\%$ in clustering, and $2\%$ in supervised classification with a few amount of labeled data.
\keywords{Deep Learning  \and Unsupervised Learning \and Disentangled Representation \and Interpretability.}

\end{abstract}
\section{Introduction}
Data-driven models with the help of Deep Learning (DL) are affecting wide areas of scientific research and the medical domain is no exception in this matter. However, in healthcare, developing a machine learning algorithm with expert level performance is important but not enough for the adoption of the algorithm when the issues of trust and explainability are not taken into consideration \cite{miotto2017deephealthcare}. Explainability of a model is approached either by 1) explicitly learning it by model design or 2) after model design such as using gradient-based localization \cite{selvaraju2017gradcam}.

Approaching explainability by model design could be facilitated in a supervised manner as in decision trees and rule-based systems or in an unsupervised manner as in Variational Autoencoder (VAE) \cite{kingma2013vae} or \acrfull{bvae}~\cite{higgins2017betavae}. In the latter, a lower dimensional representation of the data is learned and utilized for analyzing the data. The rest of the paper discusses this type of explainability. Deep learning models extract features from data in order to represent it in a compressed high-level representation that suits the application. The quality of this representation is crucial for the model performance and it is argued that disentangled representations would be helpful for having better control and interpretability over the data \cite{bengio2013representation,higgins2017betavae}. A disentangled representation can be defined as a representation where one latent unit represents one generative factor of variation in the data while being invariant to other generative factors~\cite{bengio2013representation}. For example, a model trained on a dataset of faces would learn disentangled latent units that represent independent ground truth generative factors such as hair color, pose, lighting or skin color. Disentangling as many explanatory factors as possible is important for a more compact, explainable, transferable, abstract representation of the data \cite{bengio2013representation}.

\begin{figure}[t]
\centering
\subfloat[h][]
    {
    \includegraphics[width=0.75\textwidth]{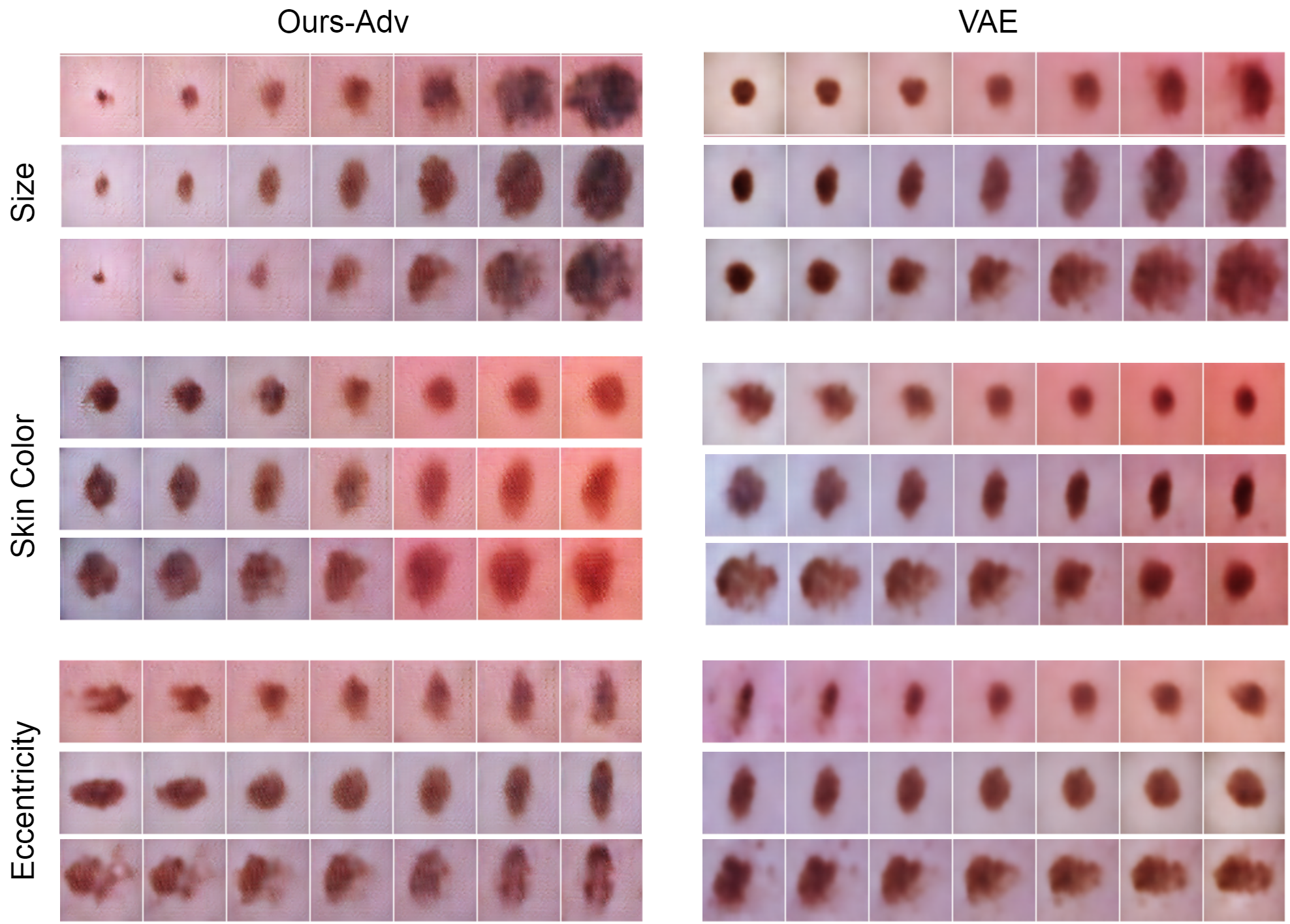} 
    \label{fig:traverse}
    }
\hfill
\subfloat[h][]
    {
    \includegraphics[width=0.20\textwidth]{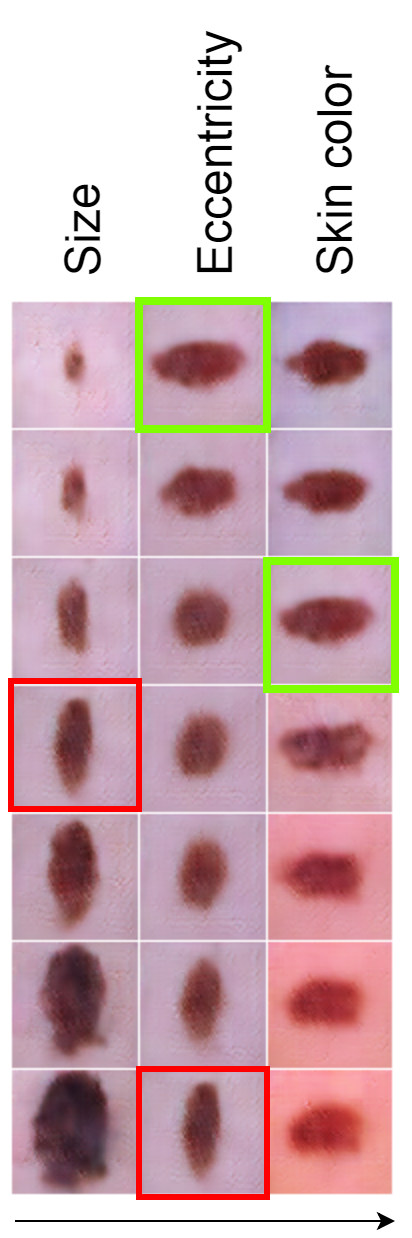} 
    \label{fig:smooth}
    }
\caption{Comparison of our model to VAE on examples for traversal over the representation components. Traversal is done between [-3, 3] (a) Examples of traversal for three images form ISIC 2018. Each row shows reconstructions of latent traversals across one latent dimension; (b) Example of a smooth transition over the manifold by changing multiple latent dimensions to go from small lesion on pale skin (top left image) to bigger horizontal lesion on red skin (bottom right image). Each column represents one dimension of change. The colored squares represent the image of the previous column from which the traversal has started on the current dimension.}
\label{fig:interpret}
\end{figure}

Most of the previous work regarding disentanglement relied on information about the number or nature of the ground truth generative factors~\cite{hinton2011transforming,kulkarni2015deepinverse}. In medical applications, the data is complex and a priori knowledge about the generative factors is mostly unavailable. Recently, multiple models for unsupervised disentangled feature learning were proposed~\cite{chen2016infogan,higgins2017betavae,kim2018factorvae,chen2018isolating}. $\beta$-VAE~\cite{higgins2017betavae} is proposed as a modification on \acrshort{vae} \cite{kingma2013vae} where the parameter $\beta$ is used to introduce more emphasis on the KL-Divergence part of the VAE objective. This enforces the posterior to match the factorized Gaussian prior which constraints the bottleneck representation to be factorized while still reconstructing the data. Higher $\beta$ values encourage more disentangled representations with a trade-off on the reconstruction error. In $\beta$-Total Correlation VAE ($\beta$-TCVAE)~\cite{chen2018isolating}, the training is focused on the total correlation part of KL term which is responsible for the factorized representation. This lowers the trade-off on the reconstruction fidelity proposed by $\beta$-VAE. \acrshort{btcvae} is validated on examples from a controlled environment with clear factors of generation. This doesn't represent the complexity of medical data and should be addressed.

\comment{
\begin{figure}[t]
\centering
\includegraphics[width=0.5\textwidth]{Images/example.png}
\caption{Examples for traversal over the representation components. Traversal is done between [-3, 3] (a) Examples of traversal for three images form ISIC 2018. Each row shows reconstructions of latent traversals across one latent dimension; (b) Example of smooth transition over the manifold by changing multiple latent dimensions to go from small lesion on pale skin (top left image) to bigger horizontal lesion on red skin (bottom right image). The colored squares represent the image of the previous row from which the traversal has started on the current dimension.} \label{example}
\end{figure}
}
\paragraph{Contributions: }In this work, we propose a framework for learning disentangled representations in medical imaging in an unsupervised manner. To our knowledge, this is the first work that analyzes the strength of unsupervised disentangled feature representations in medical imaging and proposes a framework that is well suited to medical applications. We propose a novel residual adversarial VAE with total correlation constraint
This enhances the fidelity of the reconstruction and captures more details that describe better the underlying generative factors.

\section{Methodology}
\label{sec:methodology}
We utilize deep generative disentangled representation learning to 
learn the distribution of a medical imaging dataset.
%represent images of a medical dataset.
We then use the learned representation to generate images while controlling some generative factors. We first show how disentanglement is approached with \acrshort{bvae} as a motivation for incorporating \acrshort{btcvae}. We then present our contributions to the disentanglement framework by utilizing adversarial loss with residual blocks to enhance the disentanglement and reduce the compromise on the reconstruction.
We hypothesize that using adversarial loss with residual blocks in a disentanglement framework would result in higher quality representations with more disentanglement in the feature space.% according to the Mutual Information Gap (MIG) metric which will be presented in Section~\ref{sec:experiments}.

Let $x_n \in \mathcal{X},n=1,...,N$ be a set of images generated by combinations of $K$ ground-truth generative factors $V=(v_1,...,v_K)$. Our aim is to build an unsupervised generative model that utilizes only the images in $\mathcal{X}$ to learn the joint distribution of the images and the set of latent generative factors $z \sim q_{\phi}(z|x)\in \mathbb{R}^d$ allowing us to have better control and interpretability of the latent space. It is worth mentioning the latent generative factors capture both disentangled and entangled factors. 
To realize our aim, we follow the concept of \acrshort{bvae} in learning a posterior distribution that could be used to generate images from $\mathcal{X}$. The posterior representation is approximated by $q_{\phi}(z|x)$. The model is built such that the generative factors $V$ are represented by the posterior bottleneck in a disentangled fashion.

In \acrshort{bvae}, implicit independence is enforced on the posterior to encourage a disentangled representation. This is done by constraining the posterior to match a prior $q(z)$. The prior is set to be an isotropic unit Gaussian ($p(z)=\mathcal{N}(0, I)$). Adding extra pressure on the posterior to match $p(z)$ constraints the capacity of the bottleneck and pushes it to be factorized~\cite{higgins2017betavae}. Thus, the objective function for \acrshort{bvae} is as follows
\begin{equation}
\begin{split}
    %arg\min_{\phi,\theta}\mathcal{L}_{\beta\text{-}vae}&=\mathcal{L}_{rec} + \mathcal{L}_{prior\_vae}\\
    arg\min_{\phi,\theta} \big[\underbrace{-\mathds{E}_{q_{\phi}(z|x)}[log p_{\theta}(x|z)]}_{\text{reconstruction loss }\mathcal{L}_{rec}} + \beta D_{KL}(q_{\phi}(z|x)||p(z))\big]
    %\mathcal{L}_{\beta\text{-}vae}= \mathcal{L}_{rec} + \beta D_{KL}(q_{\phi}(z|x)||p(z))
    \label{eq:bvae}
\end{split}
\end{equation}
where $\theta$ and $\phi$ are trainable weights of encoder and decoder respectively, $D_{KL}$ is the Kullback\text{-}Leibler divergence. When $\beta=1$, we get the original VAE loss \cite{kingma2013vae}. For disentanglement, values of $\beta>1$ are typically chosen. Using this formula enhances the disentanglement at the cost of reconstruction fidelity. It is suggested by~\cite{chen2018isolating} that the total correlation term within $D_{KL}$ is responsible for the factorized representation. Hence, focusing the training on the total correlation would result in better disentanglement while having less effect on the reconstruction. The objective function changes such as $D_{KL}$ is decomposed and $\beta$ is now only multiplied by the total correlation term as follows
\begin{equation}
\begin{split}
    %\mathds{E}_{q_{\phi}(z|x)}[log p_{\theta}(x|z)]
    %\mathcal{L}_{\beta\text{-}tcvae}
    arg\min_{\phi,\theta} \big[-\mathds{E}_{q_{\phi}(z|x)}[log p_{\theta}(x|z)] +\qquad\qquad\qquad\qquad\qquad\qquad \qquad\qquad\\
                                     \underbrace{I_q(z,x)+ \beta D_{KL}\big(q_{\phi}(z)||\prod_{j}q_{\phi}(z_j)\big) + \sum_{j}D_{KL}\big(q_{\phi}(z_j)||p(z_j)\big)}_{D_{KL}(q_{\phi}(z|x)||p(z))\text{ decomposition } (\mathcal{L}_{prior})}\big]
    \label{eq:tcvae2}
\end{split}
\end{equation}
The second term $I_q(z,x)$ is the mutual information between the data and the latent variable. Penalizing this term reduces the amount of information related to $x$ that are represented in $z$. Which in turn could decrease the reconstruction performance. The third term $D_{KL}(q(z)||\prod_{j}q(z_j))$ is the total correlation (TC) which is a generalization of mutual information to more than two variables. Penalizing TC forces independence in the represented factors. The last term is referred to as dimension-wise KL and is applied on individual latent dimensions. We use \acrshort{btcvae} for its good results on disentanglement on various datasets while having better reconstruction that other disentanglement models and for the parameter-less approximation of $q(z)$.
For more details about the $D_{KL}$ decomposition and the approximation of $q(z)$ the reader is referred to~\cite{chen2018isolating}.

To enhance the fidelity of the reconstructions and improve the generative factors captured by $z$, we add a discriminator network on top of \acrshort{btcvae} model. The discriminator is trained to decide whether an input image is generated synthetically or sampled from the real data distribution. We employ adversarial loss scheme for the training. The discriminator in this scenario has to learn implicitly a rich similarity metric based on features extracted from the images rather than relying only on pixel-wise similarity. This does not only improve generated images visually, but also learns a richer representation in the code $z$~\cite{larsen2015vaegan}. This is because the pixel-wise loss acts as a \textit{content} loss while the discriminator loss acts as a \textit{style} loss~\cite{gatys2015styletransfer}. 
Moreover, we incorporate residual blocks rather than convolutional layers applied in \cite{chen2018isolating}. This is because residual blocks have shown a better flow of the gradients. This limits the problems related to vanishing/exploding gradients \cite{he2016resnetv2} and is being used in
state-of-the-art Generative Adversarial Nets (GANs) literature \cite{zhang2018self} for more stable training.
We denote $Dis(.;\psi)$ to the discriminator network described by trainable parameters $\psi$, $x$ is a real image sampled from $p(x)$ and $\hat{x}$ is the reconstructed image from $p_{\theta}(x|z)$. The final objective is
\begin{equation}
\begin{split}
    %\mathds{E}_{q_{\phi}(z|x)}[log p_{\theta}(x|z)]
    \text{arg}\min_{\phi,\theta}\left[\mathcal{L}_{gen}\right]=
    \text{arg}\min_{\phi,\theta}\left[\mathcal{L}_{rec} + \mathcal{L}_{prior} - log(Dis(\hat{x}))\right]\\
    \text{arg}\min_{\psi}\left[\mathcal{L}_{disc}\right]=
    \text{arg}\min_{\psi}[-log(Dis(x))-log(1-Dis(\hat{x}))]
    \label{eq:gen}
\end{split}
\end{equation}
The model is trained by alternating between $\mathcal{L}_{gen}$ and $\mathcal{L}_{disc}$ optimization. We use pixel-wise $l_2$-distance between $x$ and $\hat{x}$ as $\mathcal{L}_{rec}$.

\section{Experiments}
\label{sec:experiments}
Experimental validation evaluates the proposed framework in two main experiments: First, we compare our proposed method disentanglement performance to state-of-the-art methods in learning both entangled and disentangled representations. We also utilize the learned representations in two use-cases, namely, unsupervised clustering and supervised classification with a few amounts of labels. 
In the second experiment, we evaluate the results visually and analyze the interpretable learned representation.

\paragraph{Dataset:}
We opt for the publicly available Skin Lesion dataset from ISIC 2018 Challenge~\cite{tschandl2018isic18} to perform our validations. To train our model, we utilize the dataset of Task 3 which consists of $10k$ RGB images with 7 types of skin lesions capturing 7 pathological generative factors. To evaluate the model against ground-truth generative factors, i.e. \textit{eccentricity, orientation, and size}, we utilize the dataset of Task 2 which consists of $2k$ images with pixel-wise segmentation. Note that all images are down-sampled to $64\times64 px$.

\paragraph{Evaluation metrics:}
To quantitatively evaluate the disentanglement quality, we report the Mutual Information Gap (MIG) metric as proposed and suggested in~\cite{chen2018isolating}. 
As opposed to the disentanglement metric in~\cite{higgins2017betavae}, MIG takes axis-alignment (one $v_k$ is captured by one $z_j$) into consideration, and it is unbiased to hyper-parameters opposite to~\cite{higgins2017betavae,kim2018factorvae}.  
MIG measures the mutual information (MI) between $z_j$ and the known generative factor $v_k$, then the difference between the two highest MIs of a generative factor is calculated, and normalized then by the entropy of $v_k$. The average MIG is then computed as 
\begin{equation}
    MIG=\frac{1}{K}\sum^{K}_{k=1}\frac{1}{H(v_k)}\bigg(I(z_{j^{(k)}},v_k)-\max_{j\neq j^{(k)}}I(z_j,v_k)\bigg),
    \label{eq:mig}
\end{equation}
where $H(\cdot)$ is the entropy, $I(\cdot,\cdot)$ is the mutual information. For our experiments, we set the generative factors as follows:
\begin{enumerate}
    \item \textbf{MIG Pathologies ($MIG_{p}$)}: The ground truth classes are used as generative factors in one vs. all fashion. For instance, $K=7$ for the Skin Lesion dataset. Each generative factor has two possible values in this scenario. 
    \item \textbf{MIG Handcrafted Factors ($MIG_{hf}$)}: In addition, we handcrafted a few generative factors which are easily visible in the image space, e.g. geometric and morphological changes. To do so, the segmentation masks given in Task 2 are utilized. The handcrafted factors are \textit{eccentricity, orientation, and size} (i.e $K=3$). Each generative factor has two possible values. %This annotation will be publicly released to pave the way for new developments in the field.
\end{enumerate}

In addition, we report the Peak signal-to-noise ratio (PSNR), and Normalized Mutual Information (NMI), and Accuracy (ACC) to evaluate the reconstruction error, clustering, and classification, respectively. 

\paragraph{Baselines:}
We compare the proposed model to two representation learning models. The first is \acrshort{vae}~\cite{kingma2013vae} model which does not take disentanglement into account explicitly. The second model is \acrshort{btcvae}~\cite{chen2018isolating} which adds constraints on the representation to disentangle the components. Further, We employ two variations of our proposed method with bottleneck residual blocks~\cite{he2016resnetv2}; 1) without the adversarial loss in Equation~\ref{eq:gen} denoted as \textit{Ours-resnet}; and 2) with the adversarial loss denoted as \textit{Ours-adv}.

\paragraph{Implementation details:}
We implement the same architecture appeared in the CelebA experiments in~\cite{chen2018isolating} for both \acrshort{vae} and \acrshort{btcvae}. For our proposed method, we replace the convolutional layers with bottleneck residual blocks for both \textit{Ours-resnet} and \textit{Ours-adv}, while the additional discriminator network in \textit{Ours-adv} has the same architecture of the encoder except for the last layer which has a single output. 
All models are trained using Adam optimizer for $100K$ iterations with a minibatch size of 256, and a learning rate of $1e-4$. $\beta$ and $d$ are set to 6 and 32, respectively. Note that we employ leakyReLU in our \textit{Ours-adv} which has been successfully applied in the adversarial training literature. 

\paragraph{Comparison with state-of-the-art:} We compare our method with the recent state-of-the-art methods by reporting the evaluation metrics (\emph{cf} Table.\ref{tbl:results}). 
We notice improvements over the \acrshort{btcvae} in terms of disentanglement with a relative improvement of $81.6\%$ and $161.8\%$ on $MIG_{p}$ and $MIG_{hf}$, respectively. For reconstruction error, it is expected that \acrshort{vae} would be superior to other models because there is no extra focus on the prior constraining part of the loss function which allows reconstruction error to optimize better. However, we notice an improvement on PSNR compared to \acrshort{btcvae} model which compromises reconstruction error for disentanglement. This experiment shows that adding the bottleneck residual blocks together with adversarial training not only improves the disentanglement, but also improves the reconstruction quality. 
 
\paragraph{Use-cases:} 
In order to show that the disentangled representation is rather capturing some meaningful generative factors, which might be relevant to the task at hand. We design two use-cases in both unsupervised and supervised paradigms. 
For the clustering use-case, we utilize the learned representations to fit a Gaussian Mixture Model (GMM) with 7 components and assign a label to each data point. NMI is then calculated between assigned labels and ground-truth labels. We report an average of 10 realizations. 
Regarding the classification use-case, we utilize the learned representations of a few amounts of labeled data to train a multi-layer perceptron (MLP) on $10\%$ of the data and evaluate it on the remaining $90\%$ of the data. 10-fold stratified cross-validation is performed. 

The model gives a relative improvement of $11.6\%$ and $2\%$ on the NMI and ACC, respectively. This could be attributed to the quality of the learned representation where features responsible for the pathologies are captured by disentanglement models as generative factors.

\begin{table}[t]
    \centering
    \caption{Comparison of various representation learning models.}
    \label{tbl:results}
    \begin{tabular}{c|c|c|c|c|c}
         \hline
          &  $MIG_p$\% & $MIG_{hf}$\% & PSNR & NMI\%& ACC\%\\
         \hline
         VAE            & 5.23 & 2.74 & \textbf{22.91}    & 9.12    & 67.88 \\
         $\beta$-TCVAE  & 6.92 & 3.53 & 20.79             & 10.66   & 68.61 \\
         \hline
         Ours-resnet & 11.61 & 5.89                   & 19.42 & 9.89 & 69.19\\
         Ours-Adv    & \textbf{12.57}& \textbf{9.24}  & 21.18 & \textbf{11.86}& \textbf{70.02}\\
         
    \end{tabular}
\end{table}

\paragraph{Interpretability:} We qualitatively examine the interpretability of the learned representations by manipulating the latent code. For instance, 
Fig.~\ref{fig:traverse} shows a comparison of the traversal between the proposed model and VAE. We notice that the dimension responsible for changing skin color has some entanglement with eccentricity and size in the case of VAE. In contrast, we can see in our proposed model that the size and eccentricity are barely changed when the skin color dimension is changed. 
For eccentricity, we notice in the case of VAE that fewer variations are captured such as the absence of the horizontal elliptic lesions that are captured with the proposed approach. 

In Fig.~\ref{fig:smooth}, we show the possibility of generating images with specific features by smoothly moving over the manifold of the representations. We show the transition of a small lesion on pale skin to a big horizontal lesion on reddish skin by changing multiple latent dimensions responsible for each feature. Having this control over the representation does not only give the ability to generate images with specific known features, but also gives an interpretable representation of the data which can be utilized in many applications.

\section{Discussion}
\label{sec:discussion}
In this paper, we introduce a novel adversarial \acrshort{vae} with a total correlation constraint to enforce disentanglement on the latent representation while preserving the reconstruction fidelity. 
The proposed framework is evaluated on skin lesions dataset and shows improvements over other state-of-the-art methods in terms of disentanglement. 
The disentangled representations learned by the proposed method has shown remarkable performance in both unsupervised clustering and supervised classification. 
We believe that our work would pave the way for other researchers to further investigate this interesting direction of research. 
One potential direction is utilizing the control over the generative factors for data augmentation.

\bibliography{literature}

\end{document}